\documentclass{article}
\usepackage{spconf,amsmath}
\usepackage{graphicx}
\usepackage{amssymb} 
\usepackage{bm}
\usepackage{color}
\usepackage{booktabs}
\usepackage{multirow}
\usepackage{rotating}
\usepackage{subfigure}
\usepackage{soul}

\title{Semantic Embedding Space for Zero-Shot Action Recognition}
\name{Xun Xu, Timothy Hospedales, Shaogang Gong}
\address{School of EECS, Queen Mary University of London, London, UK}
\begin{document}
%
\maketitle
\begin{abstract}
The number of categories for action recognition is growing rapidly. It is thus becoming increasingly hard to collect sufficient training data to learn conventional models for each category. This issue may be ameliorated by the increasingly popular ``zero-shot learning'' (ZSL) paradigm. In this framework a mapping is constructed between visual features and a human interpretable semantic description of each category, allowing categories to be recognised in the absence of any training data. Existing ZSL studies focus primarily on image data, and attribute-based semantic representations. In this paper, we address zero-shot recognition in contemporary video action recognition tasks, using semantic word vector space as the common space to embed videos and category labels. This is more challenging because the mapping between the semantic space and space-time features of videos containing complex actions is more complex and harder to learn. We demonstrate that a simple self-training and data augmentation strategy can significantly improve the efficacy of this mapping. Experiments on human action datasets including HMDB51 and UCF101 demonstrate that our approach achieves the state-of-the-art zero-shot action recognition performance.
\end{abstract}

\vspace{-0.1cm}
\section{Introduction}
\label{sec:intro}

The number and complexity of action categories of interest to be recognised in videos is growing rapidly. A consequence of the growing complexity of actions to be recognised is that more training data per category is required to learn sufficiently strong models for complex actions. Meanwhile, the growing number of categories means that it will become increasingly difficult to collect sufficient annotated training data for each. Moreover the annotation of space-time segments of video to train action recognition is more difficult and costly than annotating static images. The ``zero-shot learning'' (ZSL) paradigm has the potential to ameliorate these issues by respectively sharing information across categories; and crucially by allowing recognisers for novel categories to be constructed based on a human description of the action, rather than an extensive collection of training data.

The ZSL paradigm is most commonly been realised by using attributes \cite{lampert2013attributeZSL} to bridge the semantic gap between low-level features (e.g., MBH or HOG) and human class descriptions. Visual to attribute classifiers are learned on an auxiliary dataset, and then novel categories are specified by a human in terms of their attributes -- thus enabling recognition in absence of training data for the new categories. With a few exceptions \cite{Liu2011,Fu2014_PAMI}, this paradigm has primarily been applied to images rather than video action recognition. 

An emerging alternative paradigm to the attribute-centric strategy to bridging the semantic gap for ZSL is that of semantic embedding spaces (SES) \cite{Socher2013,Fu2014a,Habibian2014}. In this case a distributed representation of text words is generated by a model such as an unsupervised neural network \cite{Mikolov2013} trained on a large text corpus. This neural network is used to map the text string of a category name into a vector space. Regressors (contrast classifiers in the attribute case) are then used to map videos into this word vector space. Zero-shot recognition is then enabled by mapping novel category instances (via regression), and novel class names (via the neural network) to this common space and performing similarity matching. The key advantage of SES over attribute-centric approaches is that new categories can be defined trivially by naming them, without the requirement to exhaustively define each class in terms of a list of attributes -- which grows non-scaleably as the breadth of classes to recognise grows. Moreover it allows information sharing across categories (via the common regressor), and can even be used to improve conventional supervised recognition if training samples are sparse \cite{Habibian2014}. 

Although SES-based ZSL is a very attractive paradigm for the mentioned reasons, it has not previously been demonstrated in zero-shot video action recognition. This is for two reasons: (i) for many classes of complex actions, the mapping from low-level features to semantic embedding space is very complex and hard to learn reliably, and (ii) a heavy burden is placed on the generalisation capability of these regressors which need to learn a single visual to semantic embedding space mapping that is general enough to cover all action categories including unseen action categories. This can be seen as the pervasive issue \cite{Fu2014a} of domain shift between the categories on which the semantic embedding is trained and the disjoint set of categories on which it is applied for zero-shot recognition.

\begin{figure*}[!hbt]
\begin{center}
\includegraphics[width=0.98\linewidth]{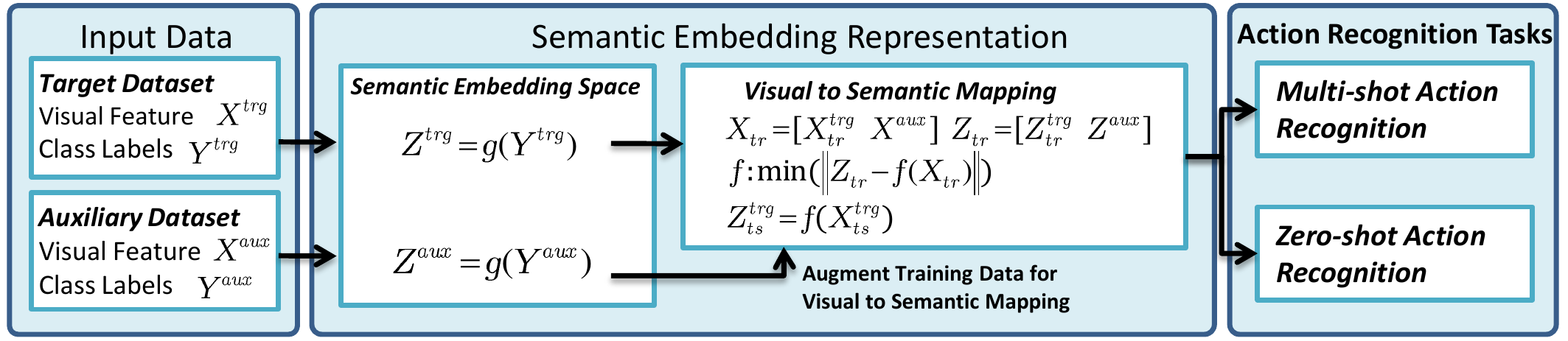}
\end{center}
\vspace{-0.4cm}
\caption{Illustration of our framework's processing pipeline. We start by exploiting word2vector $g(\cdot)$ to project textual labels $Y$ into semantic embedding space $Z$. Then we learn a Support Vector Regression $f(\cdot)$ to map low-level visual features $X$ into the semantic space$Z$. By augmenting target training data $X_{tr}^{trg}$ with auxiliary data $X_{tr}^{aux}$ we can achieve the state-of-the-art performance for Zero-shot Learning and competitive performance on Multi-shot Learning}
\label{fig:Pipeline}
\vspace{-0.3cm}
\end{figure*}

In this paper we show how to use simple data augmentation and self-training strategies to ameliorate these issues and achieve state of the art ZSL performance on contemporary video action datasets, HMDB51 and UCF101. Our framework also achieves action recognition accuracy comparable to the state of the art in the conventional supervised settings. The processing pipeline of our framework is illustrated in Fig.\ref{fig:Pipeline}.

 \vspace{-0.2cm}
\section{Methodology}
\subsection{Semantic Embedding Space}\label{subsec:semanticembedding}
To formalise the problem, we have a task $T=\{X,Y\}$ where $X=\{x_{i}\}_{i=1\cdots n}$ is the set of $d_x$ dimensional low-level space-temporal visual feature representations (e.g., MBH and HOG) of $n$ videos, including $n_{tr}$ and $n_{ts}$ training and testing samples. $Y=\{y_{i}\}_{i=1\cdots n}$  is the class names/labels of each instance (e.g. ``brush hair" and ``handwalk"). We want to establish a semantic embedding space $Z$ to connect the low-level visual features and class labels. In particular we use the word2vector neural network $\cite{Mikolov2013}$ trained on a 100 billion word corpus to realise a mapping $g:y\to z$ that produces a  unique $d_z$ dimensional  encoding of each word in the english dictionary and thus any class name of interest in $Y$. For multi-word class names, such as ``brush hair" or ``ride horse" we generate a single vector $z$ by averaging the unique words $\{y_j\}_{j=1\cdots N}$ in the description \cite{milajevs2014sentenceComposition}: $z=\frac{1}{N}\cdot\sum_{j=1}^{N}g(y_j)$.

\subsection{Visual to Semantic Mapping}\label{subsec:regression}

In order to map videos into the semantic embedding space constructed above, we train a regression model $f: x\to z$ from $d_x $ dimensional low-level space-time visual feature space to the $d_z$ dimensional embedding space. The regression is trained using training instances $X_{tr}=\{x_i\}_{i=1\cdots n_{tr}}$ and the corresponding embedding $Z_{tr}=g(Y_{tr})$ of the instance class name $y$ as the target value. Various methods have previously been used for this task including linear support vector regression (SVR) \cite{Fu2014a} and more complex multi-layer neural networks \cite{Socher2013}. Considering the trade-off between accuracy and complexity, we choose non-linear SVR with RBF-$\chi^2$ kernel defined by:

\begin{equation}\label{eq:rbfchi2_kernel}
K(x_i,x_j)=exp(-\gamma \cdot \mathcal{D}(x_i,x_j))
\vspace{-0.0cm}
\end{equation}

\noindent where $\mathcal{D}(x_i,x_j)$ is the $\chi^2$ distance between histogram based representation $x_i$ and $x_j$ \cite{Laptev2008}. This kernel is effective for histogram-based low-level space-time feature representations \cite{Wang2013} that we  use. 

\subsection{Multi-shot and Zero-shot Learning}\label{sect:MulZeroLearning}
Distances in semantic embedding spaces have been shown to be best measured using the cosine metric  \cite{Mikolov2013,Fu2014a}. Thus we normalise each data point in this space with L2, making euclidean distance comparisons effectively correspond to cosine distance $d(z_i,z_j)=1-\frac{<z_i,z_j>}{\|z_i\|\cdot\|z_j\|}$ in this space.

\noindent\textbf{Multi-shot learning}\quad
For conventional multi-shot learning, we map all data instances $X$ into the semantic space using projection $Z=f(X)$, and then simply train SVM classifiers with RBF kernel using the L2 normalised projections $f(X)$ as data.

\noindent\textbf{Zero-shot learning}\quad
For zero-shot learning, test instances and classes $X^*$ and $Y^*$ are disjoint from training classes. I.e., no instances of test classes $Y^*$ occur in training data $Y$. For each unique test category $y^*\in Y^*$, we obtain its semantic space projection $g(y^*)$. Then the embedding $f(x^*)$ of each test instance $X^*$ is generated via support vector regressor $f$ as described earlier. To classify instances $x^*$ of novel categories, we use nearest neighbour matching:

\begin{equation}\label{eq:zsl_nn}
\hat{y}^* = \arg\min_{y^*\in Y^*}\|f(x^*) - g(y^*)\|
\end{equation}

The projections $g(y^*)$ can be seen as class \emph{prototypes} in the semantic space. Data instances $f(x^*)$ can be directly matched against prototypes in this common space.

\noindent\textbf{Self-training for domain adaptation}\quad
The change in statistics induced by the disjointness of the training and zero-shot testing categories $Y$ and $Y^*$ means that regressor $f$ trained on $X$ will not be well adapted at zero-shot test time for $X^*$, and thus perform poorly \cite{Fu2014a}. To ameliorate this domain shift, we apply transductive self-training (Eq.~\ref{eq:self_train}) to adjust unseen class prototypes to be more comparable to the projected data points. For each category prototype $g(y^*)$ we search for the $K$ nearest neighbours among the unlabelled test instance projections, and re-define the adapted prototype $\tilde{g}(y^*)$ as the average of the $K$ neighbours. Thus if $NN_K(g(y^*))$ denotes the set of K nearest neighbours of $g(y^*)$, we have:
\begin{equation}\label{eq:self_train}
\tilde{g}(y^*) := \frac{1}{K}\sum_{f(x^*)\in NN_K(g(y^*))}^{K} f(x^*)
\vspace{-0.1cm}
\end{equation}
The adapted prototypes $\tilde{g}(y^*)$ are now more directly comparable with the test data for matching using Eq.~(\ref{eq:zsl_nn}).

\subsection{Data Augmentation}\label{sec:augment}
The approach outlined so far relies heavily on the efficacy of the low-level feature to semantic space mapping $f(x)$. As discussed, the mapping is hard to learn well because: (i) actions are visually complex and ambiguous, and (ii) even a mapping well learned for training categories may not generalise well to testing categories as required by ZSL, because the volume of training data is small compared to the complexity of a general visual to semantic space mapping. The self-training mechanism above addresses the latter to some extent.

Another way to further mitigate both of these problems is via augmentation with any available auxiliary dataset $T^{aux}=\{X^{aux},Y^{aux}\}$, which need not contain classes in common with the target dataset $T^{trg}$. This will provide more data to learn a better and more generalisable regressor $z=f(x)$. The auxiliary dataset class names $Y^{aux}$ are also projected into the embedding space with the neural network $g(Y^{aux})$. The auxiliary instances $X^{aux}$ are aggregated with the target training data $X_{tr}=[X^{trg}_{tr}, X^{aux}], Z_{tr}=[g(Y^{trg}_{tr}), g(Y^{aux})]$ and together to train the regressor $f$. Although the auxiliary data is disjoint from both the target training or zero-shot classes, it helps to both better learn the complex visual-semantic space mapping and to learn a more generalisable mapping that better applies to  the held-out zero-shot classes.

\vspace{-0.3cm}
\section{Experiments}

\noindent\textbf{Datasets:}\quad Experiments are performed on HMDB51~\cite{Kuehne2011} and UCF101~\cite{Soomro2012}, two of the largest and most challenging action recognition datasets available. HMDB51 has 6766 videos with 51 categories of actions. UCF101 has 13320 videos with 101 categories of actions. 

\noindent\textbf{Visual Feature Encoding:}\quad For each video we extract dense trajectory descriptors using \cite{Wang2013} and encode Bag of Words features. We first compute dense trajectory descriptors (DenseTrajectory, HOG, HOF and MBH) then we randomly sample 10,000 descriptors from all videos and learn the BoW codebook with K-means using K=4000. Thus $d_x=4000$.

\noindent\textbf{Semantic Embedding Space:}\quad  We adopted the skip-gram neural network model \cite{Mikolov2013} trained on the Google News dataset (about 100 billion words). This neural network can then encode any of approximately 3 million unique worlds as a $d_z=300$ dimension vector. 

\noindent\textbf{Visual to Semantic Mapping:}\quad The SVR from visual feature $X$ to semantic space $Z$ is learned from training data with RBF-$\chi^2$ kernel. The $\gamma$ parameter for the kernel is set as $\gamma=\frac{1}{\frac{1}{N}\cdot \sum_{i,j} \mathcal{D}(x_i,x_j)}$ where $\mathcal{D}(x_i,x_j)$ is the $\chi^2$ distance function. Slack parameter C  for SVR was set to 2.

\vspace{-0.1cm}
\subsection{Zero-shot Learning}\label{sect:exp_ZSL}

\noindent\textbf{Data Split:}\quad Because there is no existing zero-shot learning evaluation protocol for HMDB51 and UCF101 action datasets, we propose our own split\footnote{The data split will be released on our website}. For each dataset, we use a 50/50 category split. Semantic space mappings are trained on the $50\%$ training categories, and the other $50\%$ are held out unseen until test time. We randomly generate 30 independent splits and take the average mean accuracy and standard deviation for fair evaluation.

\noindent\textbf{Alternative Approaches:}\quad We compare our model with 3 alternatives: (1) Random Guess - A lower bound that randomly guesses the class label of unseen test samples. (2) Attribute Based - the classic Direct Attribute Prediction (DAP) \cite{lampert2013attributeZSL} zero-shot recognition strategy.  (3) Attribute Based - Indirect Attribute Prediction (IAP) \cite{lampert2013attributeZSL}.
Note that because attribute annotations are only available for UCF dataset, the two attribute methods are only tested on UCF. (4) Vanilla semantic word vector embedding with Nearest Neighbour (NN) - This is the simplest variant of ZSL using the same embedding space as our model. Projection $f$ is learned as for our model, and then NN matching is applied to classify test instances using the unseen class prototypes. (5) Our zero-shot learning approach (Sec.~\ref{sect:MulZeroLearning}), including both Nearest Neighbour + Self-Training (NN+ST). (Investigation of Data Augmentation is in the following Sec.~\ref{sec:dataAug}.) 

The results are presented in Tab.~\ref{tab:ZeroshotAcc}. All methods are much better than random chance, demonstrating successful ZSL. A direct application of the embedding space (NN) is reasonable, suggesting that the semantic space is effective as a representation: Videos are successfully mapped near to the correct prototypes in the semantic space. Although NN is not clearly better than the  attribute-based approaches \cite{lampert2013attributeZSL}, it does not require the latter's extensive and costly attribute annotation. Finally, our self-training approach  performs best, suggesting that our strategy ameliorates some of the domain-shift between training and testing categories compared to vanilla NN.

\begin{table}[h]
  \centering
  \caption{Zero-shot action recognition performance (average \% accuracy $\pm$ standard deviation).}
    \begin{tabular}{lrr}
    \toprule
    Method & \multicolumn{1}{c}{HMDB51} & \multicolumn{1}{c}{UCF101} \\
    \midrule
    \multicolumn{1}{l}{Random Guess} & \multicolumn{1}{c}{$4.0$} & \multicolumn{1}{c}{$2.0$} \\
    \multicolumn{1}{l}{DAP\cite{lampert2013attributeZSL}} & \multicolumn{1}{c}{--} & \multicolumn{1}{c}{$14.3 \pm 1.9$} \\
    \multicolumn{1}{l}{IAP\cite{lampert2013attributeZSL}} & \multicolumn{1}{c}{--} & \multicolumn{1}{c}{$12.8 \pm 2.0$} \\
    \multicolumn{1}{l}{NN} & \multicolumn{1}{c}{$13.0 \pm 2.7$} &  \multicolumn{1}{c}{$10.9 \pm 1.5$}\\
    \multicolumn{1}{l}{NN + ST} & \multicolumn{1}{c}{\textbf{$15.0 \pm 3.0$}} &  \multicolumn{1}{c}{\textbf{$15.8 \pm 2.3$}}\\
    \multicolumn{1}{l}{NN + Aux} & \multicolumn{1}{c}{${18.0 \pm 3.0}$} &  \multicolumn{1}{c}{\textbf{${12.7 \pm 1.6}$}}\\
    \multicolumn{1}{l}{NN + ST + Aux} & \multicolumn{1}{c}{$\mathbf{21.2 \pm 3.0}$} &  \multicolumn{1}{c}{\textbf{$\mathbf{18.6 \pm 2.2}$}}\\
    \bottomrule
    \end{tabular}%
  \label{tab:ZeroshotAcc}%
  \vspace{-0.4cm}
\end{table}%

\vspace{-0.1cm}
\subsection{Zero-shot Learning with Data Augmentation}\label{sec:dataAug}
Semantic embedding space as an intermediate representation enables exploiting multiple datasets to improve the projection $f$ as explained in Sec~\ref{sec:augment}. We next investigate the effect of data augmentation across HMDB51 and UCF101.

\noindent\textbf{Zero-shot Learning with Data Augmentation:}\quad We follow the same zero-shot learning protocol as Sec.~\ref{sect:exp_ZSL}, but augment the HMDB51 regressor training with data from UCF101 and vice versa. { The performance of only data augmentation without self-training (NN+Aux) and our full model including both self-training and data augmentation (NN+ST+Aux) are shown in }Tab.~\ref{tab:ZeroshotAcc}. { Overall the  both strategies (NN+Aux and NN+ST+Aux) significantly outperform their respective baselines (NN and NN+ST) and the full model clearly beats the classic attribute-based approaches (DAP and IAP).} 
 This is attributed to learning a more accurate and generalisable regressor for mapping videos into the semantic space for classification.  Note that NN roughly corresponds to the embedding space strategy of \cite{Socher2013} and \cite{Frome2013}. 

\begin{figure}
\subfigure[Regression trained on target seen data alone.]{\includegraphics[width=0.48\linewidth]{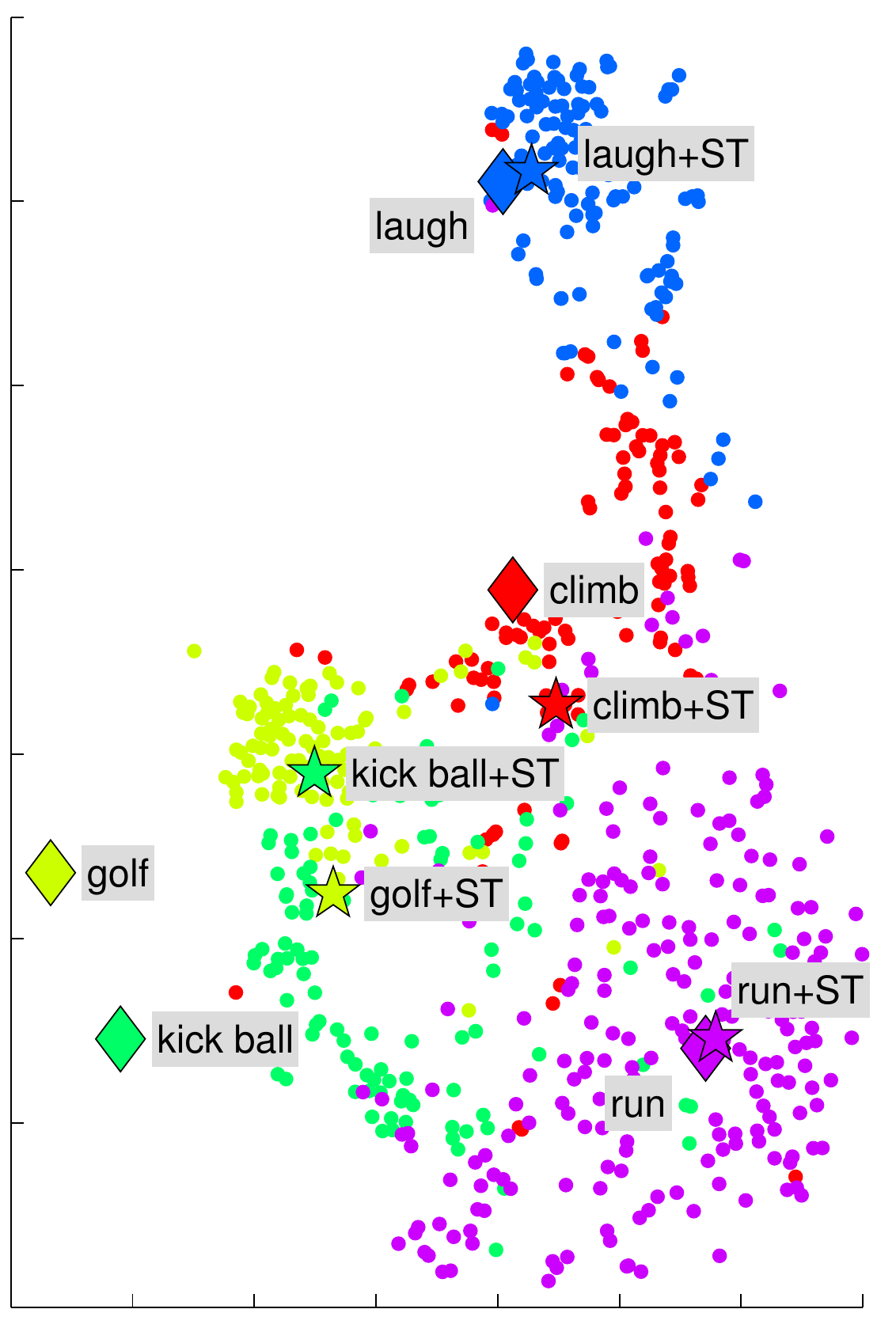}}
\subfigure[Regression trained on target data augmented with auxiliary data.]{\includegraphics[width=0.48\linewidth]{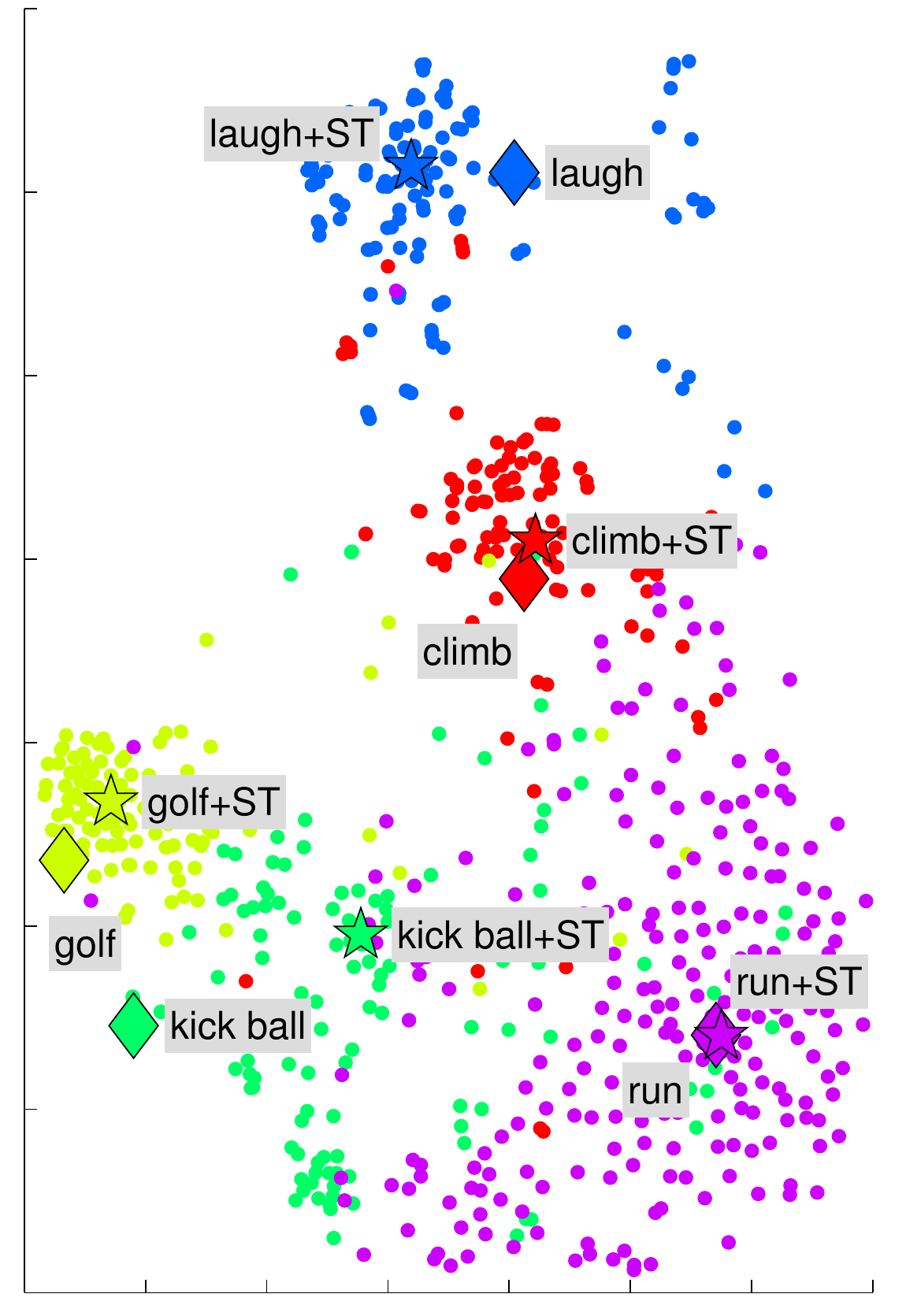}}
\vspace{-0.4cm}
\caption{A qualitative illustration of ZSL with semantic space representation: self-training and data augmentation.}
\vspace{-0.4cm}
\label{fig:TSNE_visualization}
\end{figure}

\noindent\textbf{Qualitative Illustration:}\quad 
We give insight into our self-training and data-augmentation contributions in Fig.~\ref{fig:TSNE_visualization}. We randomly sample 5 unseen classes from HMDB and project all samples from these classes into the semantic space by (a) regression trained on target seen class data alone; (b) regression trained on target seen data augmented with auxiliary (UCF101) data. The results are visualised in 2D with t-SNE\cite{Maaten2008}. Data instances are shown as dots,  prototypes as diamonds, and self-training adapted prototypes as stars. Colours indicate category. 

There are two main observations: (i) Comparing Fig.~\ref{fig:TSNE_visualization}(a) and (b), we can see that regression trained without auxiliary data yields a less accurate projection of unseen data, as instances are projected further from the prototypes: reducing NN matching accuracy. (ii) Self-training is effective as the adapted prototypes (stars) are  closer to the center of the corresponding samples (dots) than the original prototypes (diamonds). These observations explain our final model's  ZSL accuracy improvement on conventional approaches.

\vspace{-0.2cm}
\subsection{Multi-shot Learning}\label{sect:MultishotLearning}

We finally validate our representation on standard supervised (multi-shot) action recognition. We use the standard data splits for both HMDB51\cite{Kuehne2011} and UCF101\cite{Soomro2012}. The action recognition accuracy is the average of each fold. 

\noindent \textbf{Alternatives:}\quad We compare our approach to: (i) the state of the art results based on low-level features \cite{Wang2013}, (ii) an alternative semantic space using attributes. To realise the latter we use Human-Labelled Attribute (HLA) \cite{Zheng2014}. We train binary SVM classifier with RBF-$\chi^2$ kernel for attribute detection and use the concatenation of attribute scores as semantic attribute space representation. A SVM  classifier with RBF kernel is trained on attribute representation to predict final labels. 

The resulting accuracies are shown in Tab.~\ref{tab:MultishotAcc}. We  observe that our semantic embedding  is comparable to the state of the art low-level feature-based classification and better than the conventional attribute-based intermediate representation. This may be due to the attribute-space being less discriminative than our semantic word space, or due to the reliance on human annotation: some annotated attributes may not be detectable, or may be detectable but not discriminative for class. 

\begin{table}[t]
  \centering
  \caption{Standard supervised action recognition mean accuracy in \% on HMDB51 and UCF101 . $^{*}$ indicates our implementation}
    \begin{tabular}{lcr}
    \toprule
     Method     & HMDB51 & UCF101 \\
    \midrule
    Low-Level Feature\cite{Wang2013} & $47.2$/$46.0^{*}$ & \multicolumn{1}{c}{$75.1$} \\
    HLA~\cite{Zheng2014} & -- & \multicolumn{1}{c}{$69.7$} \\
    Ours  & $44.5$  & \multicolumn{1}{c}{$73.7$} \\
    \bottomrule
    \end{tabular}%
  \label{tab:MultishotAcc}%
  \vspace{-0.3cm}
\end{table}%

\vspace{-0.1cm}
\section{Conclusion}
In this paper we investigated semantic-embedding space representations for video action recognition for the first time. This representation enables projecting visual instances and category prototypes into the same space for zero-shot recognition, however it possess serious challenges of projection complexity and generalisation across domain-shift. We show that simple self-training and data augmentation strategies can address these challenges and achieve the state of the art results for zero-shot action recognition in video.

\bibliographystyle{IEEEbib}
\bibliography{icip_ref}

\end{document}